\documentclass[runningheads]{llncs}
\usepackage[T1]{fontenc}
\usepackage{xcolor}
\usepackage{graphicx}
\usepackage{amsmath}
\usepackage{amssymb}
\usepackage{booktabs}
\usepackage{multirow}
\usepackage{tabularx}
\usepackage{tabularray}
\usepackage{svg}
\usepackage{adjustbox}  

\usepackage{subcaption} 
\usepackage{caption}  

\usepackage[pagebackref,breaklinks,colorlinks]{hyperref}
\begin{document}
\title{Vision Transformers for Kidney Stone Image Classification: A Comparative Study with CNNs}%
\titlerunning{Kidney Stone Recognition with ViT}


\author{Ivan Reyes-Amezcua\inst{1}\and
Francisco Lopez-Tiro\inst{2,3}\and
Clement Larose \inst{3,4} \and \\ 
Andres Mendez-Vazquez\inst{1} \and
Gilberto Ochoa-Ruiz\inst{2}\and
Christian Daul\inst{3}
}
\institute{CINVESTAV, Guadalajara, Mexico\\
\email{ivan.reyes@cinvestav.mx, andres.mendez@cinvestav.mx} \and
Tecnológico de Monterrey, School of Engineering and Sciences, Mexico\\
\email{franciscolt@tec.mx, gilberto.ochoa@tec.mx} \and
CRAN (UMR 7039), Université de Lorraine, CNRS, Vandœuvre-les-Nancy, France \\
\email{christian.daul@univ-lorraine.fr} 
 \and
CHRU de Nancy, Service d’urologie de Brabois, Vand{\oe}uvre-l\`es-Nancy, France
\email{C.LAROSE@chru-nancy.fr}}
\authorrunning{I. Reyes et al.}
\maketitle              
%
\begin{abstract}
Kidney stone classification from endoscopic images is critical for personalized treatment and recurrence prevention. While convolutional neural networks (CNNs) have shown promise in this task, their limited ability to capture long-range dependencies can hinder performance under variable imaging conditions. This study presents a comparative analysis between Vision Transformers (ViTs) and CNN-based models, evaluating their performance on two ex vivo datasets comprising CCD camera and flexible ureteroscope images. The ViT-base model pretrained on ImageNet-21k consistently outperformed a ResNet50 baseline across multiple imaging conditions. For instance, in the most visually complex subset (Section patches from endoscopic images), the ViT model achieved 95.2\% accuracy and 95.1\% F1-score, compared to 64.5\% and 59.3\% with ResNet50. In the mixed-view subset from CCD-camera images, ViT reached 87.1\% accuracy versus 78.4\% with CNN. These improvements extend across precision and recall as well. The results demonstrate that ViT-based architectures provide superior classification performance and offer a scalable alternative to conventional CNNs for kidney stone image analysis.
\end{abstract}

\section{Introduction}

\subsection{Clinical context}
Kidney stone disease is a prevalent and recurrent health problem that affects approximately 10\% of individuals in industrialized nations, with a recurrence rate of 40\% in five years \cite{zeng2019retrospective}. 
Therefore, determining the specific type of kidney stone is crucial to prescribe an appropriate treatment and reduce recurrences. For this purpose, different techniques have been developed in the clinical practice to identify the types of kidney stones, such as the Morpho-Constitutional Analysis (MCA)  \cite{daudon2004clinical} and Endoscopic Stone Recognition (ESR) \cite{estrade2017should}.

\begin{figure*}[!ht]
\centering
\includegraphics[width=\linewidth]{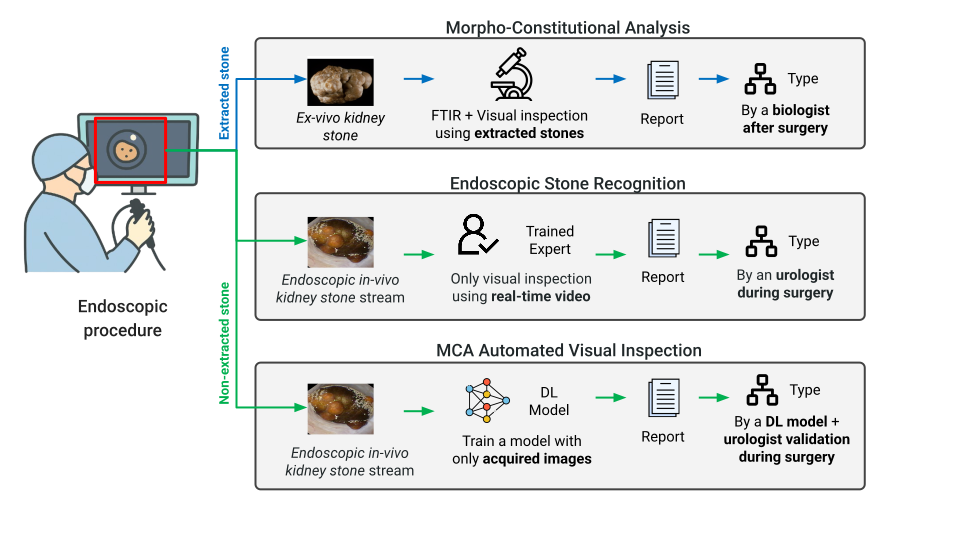}
\caption{The MCA is the standard guideline for the identification of kidney stones (top row) \cite{daudon2016comprehensive}). MCA performs a visual inspection and is complemented by a biochemical analysis on extracted kidney stones (post-surgery). Endoscopic Stone Recognition (ESR, \cite{estrade2013place}) is a technique to perform in-vivo kidney stone identification during surgery using only the information displayed on the screen (middle row). MCA automated visual inspection is a method based on computer vision techniques and machine learning to classify in-vivo/ex-vivo endoscope images (bottom row).}
\label{fig:identification}
\end{figure*}

On one hand, MCA is considered the \textit{gold standard} for identifying and classifying up to 21 types of extracted kidney stones (ex vivo) \cite{corrales2021classification} (Fig. \ref{fig:identification}, top row). This procedure involves a two-complementary analysis of kidney stone fragments. 
First, a visual inspection of the superficial characteristics (color and texture) of the external (surface) and internal (cross-sectional view) aspects of the kidney stone is performed. Visual inspection is carried out using a lens that magnifies details such as layers, core, and textures \cite{corrales2021classification}. Then, to complement the study, a Fourier Transform Infrared Spectroscopy (FTIR)  analysis is performed which unequivocally determines the type of kidney stone based on its biochemical composition  \cite{khan2018fourier}. Finally, the information obtained from the visual analysis and the FTIR analysis is combined into a report that the urologist uses to determine the treatment.
However, performing MCA requires specialized personnel and equipment for both visual inspection and FTIR analysis. Although the procedure itself can be completed within minutes, many hospitals lack the necessary resources, necessitating the shipment of samples to external laboratories. In such cases, the determination of the kidney stone type depends on the availability and processing time of these specialized centers, which can delay diagnosis and treatment by several weeks or even months.

Consequently, ESR offers a promising in vivo alternative for identifying common kidney stone types during ureteroscopy \cite{estrade2017should} (Fig. \ref{fig:identification}, middle row). It relies solely on visual features, avoiding the need for FTIR analysis. A recent study \cite{estrade2021toward} found a strong correlation between expert endoscopic assessments and MCA results.

In vivo visual recognition of the type of stone in endoscopic images could save time, as the fragments could be pulverized instead of extracting them, thus avoiding the need for MCA. However, it has been observed that laser fragmentation can alter the composition of the stones, which may bias the analyses \cite{taratkin2021lasers}. Additionally, few urologists are trained to perform this recognition efficiently, and the task is heavily operator dependent.

\subsection{MCA Automated visual inspection and Paper Contribution}

Innovative approaches based on Deep Learning (DL) techniques have been developed to automate and optimize the identification of kidney stones \cite{lopez2021assessing,lopez2024vivo}. This automation will support urologists in making immediate decisions during ureteroscopy procedures \cite{ali2022we}.
Early implementations of MCA automated visual inspection (Fig. \ref{fig:identification}) have predominantly utilized convolutional neural networks (CNN) with transfer learning from ImageNet-pre-trained models \cite{lopez2023boosting}. Although, CNNs have achieved notable success in various image classification tasks, their application in endoscopic kidney stone analysis faces significant challenges. These include insufficient feature extraction capabilities to capture fine-grained differences and, in addition, a high susceptibility to variations inherent to endoscopic imaging. Such artifacts include inconsistent lighting, the presence of bodily fluids and motion-induced artifacts rendering CNN models less useful \cite{yin2024application}.
ViTs have recently emerged as a compelling alternative to basic CNNs demonstrating superior performance across multiple medical imaging tasks \cite{halder2024implementing}. Unlike CNNs, which rely on local receptive fields, ViTs employ self-attention mechanisms to capture global patch dependencies within an image, facilitating the learning of complex and comprehensive visual representations \cite{halder2024implementing}. This global context modeling is particularly advantageous in medical imaging, where understanding the spatial relationships between anatomical structures is crucial for accurate diagnosis. ViTs  also offer better interpretability through attention-based visualization and require less data for training because their self-attention mechanism efficiently captures global contextual relationships—focusing on the most informative features even with limited training examples \cite{habib2024optimizing}.

\begin{figure*}[!ht]
\centering
\includegraphics[width=\linewidth]{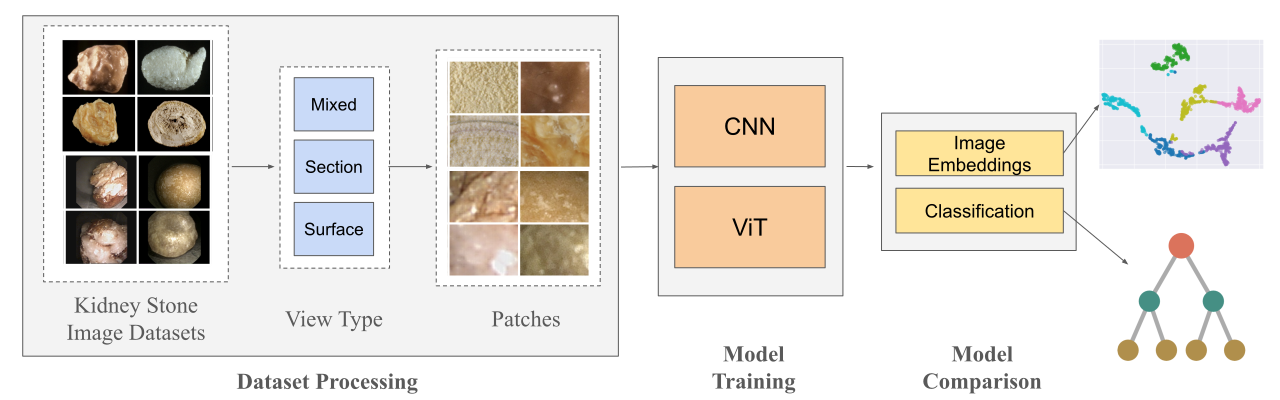}
\caption{Overview of the proposed pipeline for kidney stone classification. The process begins with dataset processing, where images are categorized by view type (Mixed, Section, Surface) and segmented into patches. Two architectures, CNN and ViT, are trained for classification. The resulting models are then compared based on classification accuracy and image embeddings, providing insights into feature representation and decision boundaries.}
\label{fig:overview}
\end{figure*}

Using the global feature extraction capabilities of ViTs, models can be developed to robustly address the challenges posed by endoscopic imaging conditions. This paper presents a contribution focused on the training, testing and comparative evaluation of ViT and CNNs for the classification of image from kidney stones. A ViT-based model is proposed to overcome limitations observed in CNN-based approaches. A comprehensive performance assessment is conducted, highlighting differences in classification accuracy and feature representation. The analysis provides insight into the advantages of transformer-based models in medical imaging tasks, with potential implications for clinical decision support systems (see Figure \ref{fig:overview}).

\subsection{Paper organization}

The remainder of this paper is structured as follows. Section \ref{sec:datasets} provides a detailed description of the two datasets used in our study. Section \ref{sec:vit_model} introduces the ViT model, outlining its architecture, key components such as patch-based processing and self-attention mechanisms, and its advantages over traditional CNN-based approaches. Section \ref{sec:training} details the training setup, whereas Section \ref{sec:results} presents a comparative analysis of our ViT-based approach against CNN architectures. This section also includes an ablation study evaluating the impact of model size and pretraining strategy on performance. Subsequently, in Section \ref{sec:feature_visualization} we carry out several feature visualization and other qualitative comparisons. Finally, Section \ref{sec:conclusion} summarizes our key findings, highlights the impact of ViT-based classification in MCA automated visual inspection, and outlines future research directions.

\section{Datasets}\label{sec:datasets}

In the experiments, two kidney stone datasets were used (see Figure \ref{tab:dataset_description}) to assess the performance of the ViT model. Images were acquired using standard CCD (Charge-Coupled Device) cameras \cite{corrales2021classification} and flexible ureteroscopes for endoscopic imaging \cite{el2022evaluation}. The key characteristics of both datasets are summarized in Table \ref{tab:dataset_description}.

\begin{table}[b!]
\caption{Detailed breakdown of the two ex-vivo kidney stone datasets used in this study. Dataset A (ex-vivo CCD-camera) \cite{corrales2021classification}, and  B (ex-vivo endoscopic) \cite{el2022evaluation}, contain kidney stone images classified by subtype, main biochemical composition, and key identifiers. Each dataset includes images captured from three perspectives: surface, section, and mixed views (surface + section). 
}
\vspace{1mm}
\label{tab:dataset_description}
\begin{tabular}{@{}cccrccc@{}}
\toprule
Dataset & Domain             & Subtype & Main composition     & Surface      & Section      & Mixed        \\ \midrule \vspace{-0.5mm}
A       & Ex-vivo CCD-Camera & Ia      & Whewellite (WW)    & 50           & 74           & 124          \\ \vspace{-0.5mm}
A       & Ex-vivo CCD-Camera & IVa1    & Carbapatite (CAR) & 18           & 18           & 36           \\ \vspace{-0.5mm}
A       & Ex-vivo CCD-Camera & IVa2    & Carbapatite (CAR2) & 36           & 18           & 54           \\ \vspace{-0.5mm}
A       & Ex-vivo CCD-Camera & IVc     & Struvite (STR)     & 25           & 19           & 44           \\ \vspace{-0.5mm}
A       & Ex-vivo CCD-Camera & IVd     & Brushite (BRU)     & 43           & 17           & 60           \\ \vspace{-0.5mm}
A       & Ex-vivo CCD-Camera & Va      & Cystine (CYS)      & 37           & 11           & 48           \\ \vspace{-0.5mm}
        &                    &         &\textbf{Total images}& \textbf{209} & \textbf{157} & \textbf{366} \\ \toprule
B       & Ex-vivo Endoscopic & Ia      & Whewellite (WW)    & 62           & 25           & 87           \\ \vspace{-0.5mm}
B       & Ex-vivo Endoscopic & IIa     & Weddellite (WD)    & 13           & 12           & 25           \\ \vspace{-0.5mm}
B       & Ex-vivo Endoscopic & IIIa    & Uric Acid (UA)     & 58           & 50           & 108          \\ \vspace{-0.5mm}
B       & Ex-vivo Endoscopic & IVc     & Struvite (STR)     & 43           & 24           & 67           \\ \vspace{-0.5mm}
B       & Ex-vivo Endoscopic & IVd     & Brushite (BRU)     & 23           & 4            & 27           \\ \vspace{-0.5mm}
B       & Ex-vivo Endoscopic & Va      & Cystine (CYS)      & 47           & 48           & 95           \\ \vspace{-0.5mm}
        &                    &         &\textbf{Total images}& \textbf{246} & \textbf{163} & \textbf{409} \\ \bottomrule
\end{tabular}
\end{table}

\subsection{Patch Extraction and Data Augmentation}

\begin{figure*}[h]
    \centering
    \subfloat[Dataset A] { 
    \label{fig:dataseta}\includegraphics[width=0.485\linewidth]{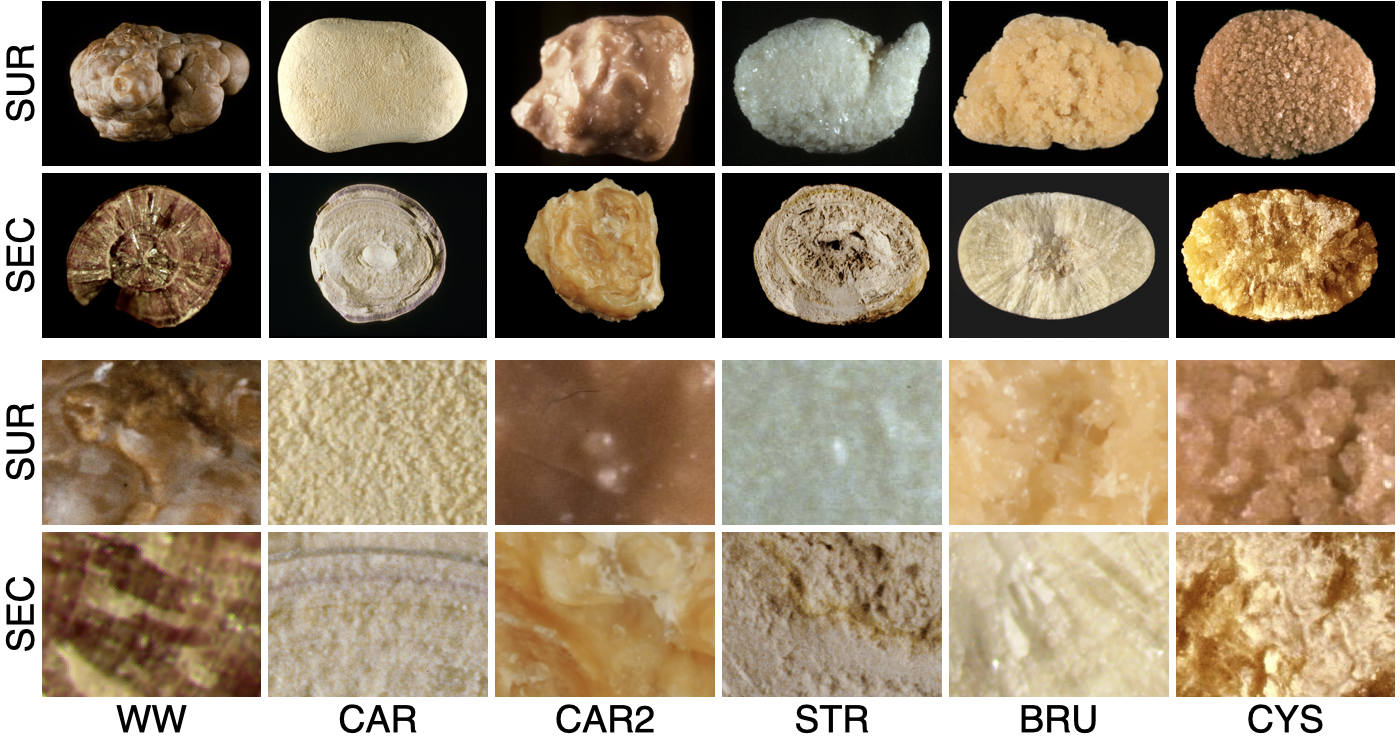}}
    \hspace{1mm}
    \subfloat[Dataset B]{
    \label{fig:datasetb}\includegraphics[width=0.485\linewidth]{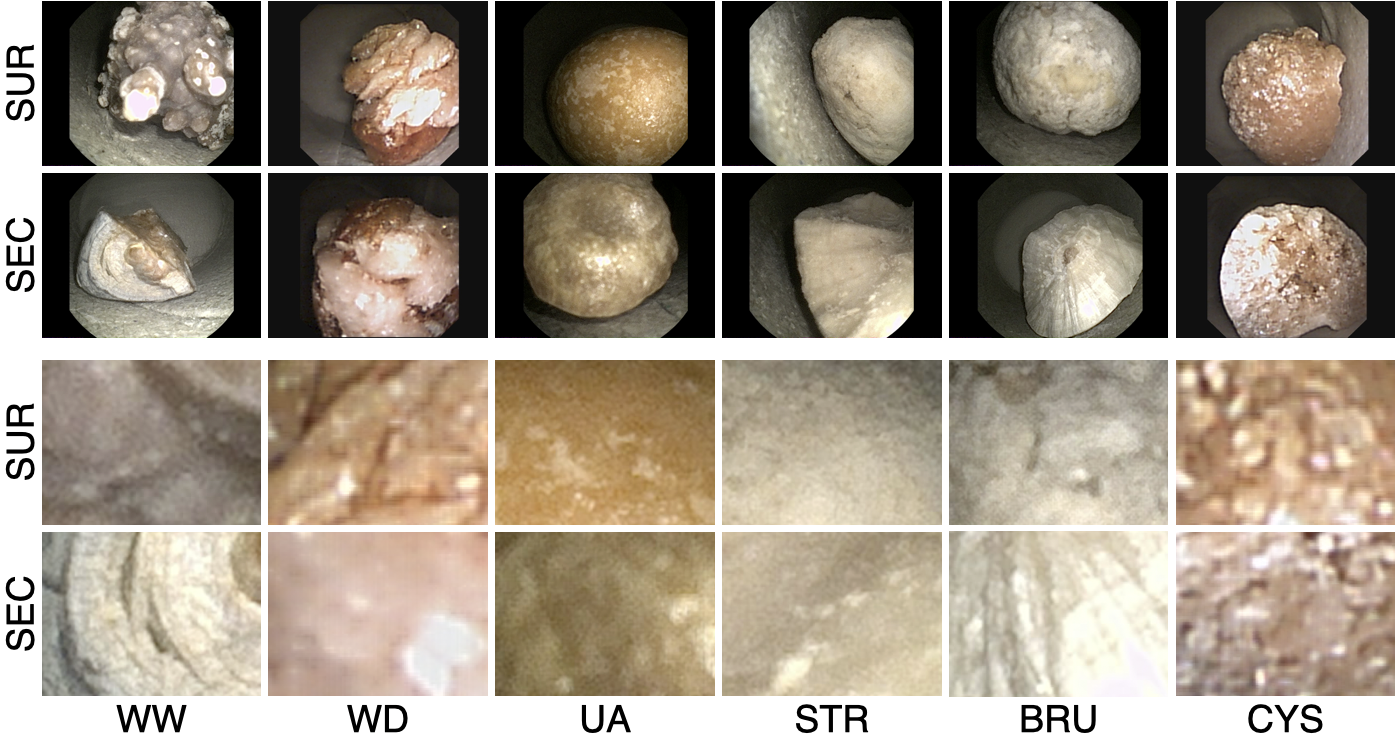}}
    \vspace{-1.5mm}
    \caption{Examples of ex-vivo kidney stone images acquired with (a) CCD camera \cite{corrales2021classification} and  (b) a flexible endoscope \cite{el2022evaluation}. The class types are defined in Table \ref{tab:dataset_description}.}
    \label{fig:dataset}
    \end{figure*}

The evaluation leverages two ex-vivo kidney stone datasets with complementary imaging modalities, see Figure \ref{fig:dataset}. Dataset A \cite{corrales2021classification} contains 366 high-resolution images captured using CCD sensors under controlled illumination, minimizing motion blur and specular reflections. In contrast, Dataset B \cite{el2022evaluation} comprises 409 endoscopic images acquired under simulated clinical conditions using a flexible ureteroscope, exhibiting realistic visual challenges such as lighting variability, motion-induced blur, and reduced spatial resolution.

All images were manually segmented under expert supervision to ensure accurate location of the kidney stones regions. According to the pre-processing protocol described in \cite{lopez2021assessing}, square patches of size $256 \times 256$ pixels were extracted from the segmented regions. This approach excluded surrounding tissue and balanced the representation of stone types while increasing the sample count. To minimize redundancy, a maximum overlap of 20 pixels was allowed between adjacent patches.

This extraction process resulted in 12,000 patches per dataset (A or B), supporting the development of a diverse and balanced training corpus. Datasets A and B include three subsets corresponding to different views: surface (SUR) as the external view, section (SEC) as the internal view, and mixed (MIX), which combines information from both (SUR + SEC). For each subset (SUR or SEC), 6000 patches were generated and organized into 4800 for training and 1200 for testing, ensuring no data leakage, while the MIX view aggregates both, yielding a total of 12,000 patches.

To standardize the input space and reduce sensitivity to illumination differences, each patch was subjected to a whitening normalization process ($w$). For each RGB channel, the mean ($\mu$) and standard deviation ($\sigma$) were calculated, and the pixel values ($I$) were normalized as follows:

\begin{equation}
I_w = \frac{I - \mu}{\sigma}
\end{equation}

This normalization ensured consistent pixel value distributions across samples, facilitating improved generalization in the presence of lighting and texture variability.

\subsection{Dataset Partitioning}

To support robust evaluation while preventing data leakage, a stratified and non-repeating partitioning strategy was applied. Each dataset was split into disjoint subsets, allocating 80\% of the patches to training and the remaining 20\% to testing, following the recommendations of \cite{lopez2024vivo}. Overlaps or repeated spatial regions between the sets were strictly avoided to ensure a clean separation.

To further enhance the robustness of the model to intra-class and acquisition variability, data augmentation techniques were applied during training. These included random geometric transformations such as rotations and horizontal flips, as well as contrast adjustments. Augmentation was particularly beneficial for Dataset B, which reflects the variability and artifacts encountered in realistic endoscopic imaging scenarios.

\section{Vision Transformer Model}\label{sec:vit_model}
Our proposed model is based on the Vision Transformer  architecture, a novel approach that leverages transformer models for image classification tasks. Unlike traditional CNN, ViTs process images by dividing them into fixed-size patches and applying a transformer encoder to capture global dependencies \cite{dosovitskiy2020image}. This methodology offers a fresh perspective on image analysis, differing from CNNs in several key aspects: 

\textit{Patch-Based Processing,} in ViTs, an input image is partitioned into non-overlapping patches, each treated as a token in a sequence. These patches are linearly embedded into vectors, allowing the model to process the image as a sequence of patches rather than a grid of pixels. This approach allows the model to capture relationships between distant parts of the image, which is challenging for CNNs due to their localized receptive fields. 

\textit{Self-Attention Mechanism,} ViTs employ a self-attention mechanism where each patch attends to every other patch in the image. This mechanism enables the model to capture long-range dependencies and contextual information across the entire image, facilitating a comprehensive understanding of complex visual patterns.

\textit{Positional Encoding,} Transformers inherently lack spatial awareness since they are designed for sequential data. To address this, ViTs incorporate learnable positional encodings to each patch embedding, preserving the spatial structure of the image and allowing the model to discern the relative positions of patches.

\textit{Classification Head,} After processing the patch sequence through the transformer encoder, ViTs utilize a fully connected layer as a classification head. This layer interprets the aggregated information from the encoder to assign a class label to the image. The design of this classification head is crucial, as it directly influences the ability of model to map the learned representations to specific categories. 

\section{Training Setup}\label{sec:training}

In the training setup, it is employed several key components to optimize the performance of our ViT  model. Specifically, we use the ViT model pre-trained on ImageNet-21k, a large-scale dataset containing 14 million images and 21,843 classes, at a resolution of $224 \times 224$ pixels. This model was introduced in  by Dosovitskiy et. al \cite{dosovitskiy2020image}. The pre-trained weights provide a rich set of visual features learned from diverse image categories, offering a strong foundation for domain-specific fine-tuning in kidney stone classification. The AdamW optimizer \cite{loshchilov2017decoupled} was used with a learning rate of 0.0001 and a batch size of 32, providing stable convergence and effective regularization. Models were trained for 30 epochs, a duration empirically chosen to ensure convergence while avoiding overfitting given the dataset size.

\begin{table}[b!]
\centering
\caption{Performance comparison of CNNs and ViTs for kidney stone classification across different datasets and patch types. Results are reported in the format: Mean ± standard deviation. CNN results correspond to ResNet50 models trained in this study, as it represents the standard architecture for kidney stone classification using CNN \cite{lopez2023boosting}. Dataset A \cite{corrales2021classification} includes ex-vivo CCD-camera images, while Dataset B \cite{el2022evaluation} comprises ex-vivo endoscopic images. MIX, SEC, and SUR denote Mixed, Section, and Surface patch types, respectively. Best scores for each metric are highlighted in bold.}
\label{tab:highlighted_comparison}
\resizebox{\linewidth}{!}{%
\begin{tblr}{
  cells = {c},
  cell{2}{1} = {r=4}{},
  cell{2}{2} = {r=2}{},
  cell{4}{2} = {r=2}{},
  cell{6}{1} = {r=4}{},
  cell{6}{2} = {r=2}{},
  cell{8}{2} = {r=2}{},
  cell{10}{1} = {r=4}{},
  cell{10}{2} = {r=2}{},
  cell{12}{2} = {r=2}{},
  hline{1,14} = {-}{0.08em},
  hline{2,6,10} = {-}{0.05em},
}
\textbf{Patch} & \textbf{Dataset} & \textbf{Model} & \textbf{Accuracy} & \textbf{F1} & \textbf{Precision} & \textbf{Recall}\\
MIX & B & CNN & 0.850±0.032 & 0.851±0.035 & 0.858±0.030 & 0.850±0.032\\
 &  & ViT & \textbf{0.901±0.011} & \textbf{0.900±0.010} & \textbf{0.907±0.009} & \textbf{0.901±0.011}\\
 & A & CNN & 0.784±0.013 & 0.781±0.013 & 0.787±0.013 & 0.784±0.013\\
 &  & ViT & \textbf{0.871±0.016} & \textbf{0.870±0.017} & \textbf{0.879±0.021} & \textbf{0.871±0.016}\\
SEC & B & CNN & 0.645±0.041 & 0.593±0.043 & 0.678±0.021 & 0.645±0.041\\
 &  & ViT & \textbf{0.952±0.026} & \textbf{0.951±0.027} & \textbf{0.956±0.022} & \textbf{0.952±0.026}\\
 & A & CNN & 0.809±0.022 & 0.805±0.023 & 0.835±0.020 & 0.809±0.022\\
 &  & ViT & \textbf{0.920±0.008} & \textbf{0.920±0.008} & \textbf{0.924±0.006} & \textbf{0.920±0.008}\\
SUR & B & CNN & 0.864±0.033 & 0.862±0.032 & 0.870±0.029 & 0.864±0.033\\
 &  & ViT & \textbf{0.887±0.014} & \textbf{0.885±0.013} & \textbf{0.896±0.012} & \textbf{0.887±0.014}\\
 & A & CNN & \textbf{0.796±0.050} & \textbf{0.798±0.046} & 0.804±0.039 & \textbf{0.796±0.050}\\
 &  & ViT & 0.792±0.017 & 0.790±0.015 & \textbf{0.811±0.018} & 0.792±0.017
\end{tblr}
}
\end{table}

\section{Results and Discussion}\label{sec:results}

\subsection{Comparison with CNN-based Approaches}

To contextualize the performance of the proposed ViT-based model, comparisons were made against a convolutional baseline using the ResNet50 architecture (Table \ref{tab:highlighted_comparison}). ResNet50 was selected as it represents the standard architecture in prior work on kidney stone classification using CNNs \cite{lopez2023boosting}, offering a well-established and reproducible benchmark. Its widespread adoption in medical image analysis, combined with a favorable trade-off between depth and computational cost, makes it a suitable reference for evaluating the effectiveness of transformer-based approaches. The results demonstrate that ViTs consistently outperform ResNet50 across multiple metrics. Specifically, ViTs achieve an overall accuracy ranging from 87.1\% to 95.2\% across different datasets and patch types, surpassing ResNet50 models, which reach accuracies between 64.5\% and 86.4\%.


These findings suggest that ViT-based models offer superior performance in the classification of kidney stone images compared to traditional CNN architectures, making them a promising alternative for clinical applications.

\subsection{Ablation Studies}

The results presented in Table \ref{tab:ablation_study} demonstrate the critical role of pretraining and model capacity in kidney stone classification. The study includes a comparative analysis of models trained from scratch (no pretraining), models pretrained on ImageNet-1k (standard weights), and those benefiting from large-scale pretraining using ImageNet-21k. Among these, ViT-base pretrained on ImageNet-21k consistently achieves the highest scores across datasets and patch subversions, confirming the effectiveness of large-scale transfer learning.

The experiments also evaluate architectural variations in both model families. For Vision Transformers, two configurations were tested: ViT-small and ViT-base, following the specifications from the original ViT paper \cite{dosovitskiy2020image}. In the case of CNNs, both ResNet50 and the deeper ResNet152 were assessed to investigate the impact of network depth. Interestingly, ResNet50 outperforms ResNet152 in most metrics, likely due to overfitting or diminishing returns from the increased depth, particularly given the limited size and variability of the medical imaging datasets. This observation underscores the importance of balancing model complexity with data characteristics.

Results further indicate that ViTs outperform CNNs, particularly in challenging views such as SEC and SUR, due to their superior ability to capture long-range dependencies and global contextual features. Smaller variants, such as ViT-small and ResNet50, generally underperform compared to their respective larger counterparts, except in cases where increased depth introduces overfitting. Overall, ViTs with strong pretraining, especially those using ImageNet-21k, demonstrate the most robust and accurate performance for kidney stone classification.

\renewcommand{\arraystretch}{1}
\setlength{\tabcolsep}{4pt}

\begin{table*}[!h]
\centering
\caption{Ablation study comparing different models, sizes, and pretraining strategies across datasets and patch subversions. Best results per group are in bold.}
\label{tab:ablation_study}

{\scriptsize
\begin{tabular}{ccccccccc}
\toprule
\textbf{Dataset} & \textbf{Subversion} & \textbf{Model} & \textbf{Size} & \textbf{Pretrained} & \textbf{Accuracy} & \textbf{F1} & \textbf{Precision} & \textbf{Recall} \\
\midrule
\multirow{9}{*}{A} & \multirow{9}{*}{MIX} 
& ResNet50 & base & ImageNet-1k & 0.784 & 0.781 & 0.787 & 0.784 \\
&& ResNet50 & base & None         & 0.631 & 0.619 & 0.637 & 0.631 \\
&& ResNet152 & base & ImageNet-1k & 0.773 & 0.765 & 0.780 & 0.773 \\
&& ResNet152 & base & None         & 0.648 & 0.648 & 0.667 & 0.648 \\
&& ViT       & base & ImageNet-1k & 0.624 & 0.621 & 0.654 & 0.624 \\
&& ViT       & base & ImageNet-21k & \textbf{0.890} & \textbf{0.889} & \textbf{0.903} & \textbf{0.889} \\
&& ViT       & base & None         & 0.209 & 0.114 & 0.266 & 0.209 \\
&& ViT       & small & ImageNet-1k & 0.623 & 0.623 & 0.666 & 0.623 \\
&& ViT       & small & None         & 0.447 & 0.435 & 0.454 & 0.447 \\
\midrule
\multirow{9}{*}{A} & \multirow{9}{*}{SEC} 
& ResNet50 & base & ImageNet-1k & 0.809 & 0.805 & 0.835 & 0.809 \\
&& ResNet50 & base & None         & 0.729 & 0.728 & 0.742 & 0.729 \\
&& ResNet152 & base & ImageNet-1k & 0.765 & 0.756 & 0.788 & 0.765 \\
&& ResNet152 & base & None         & 0.693 & 0.693 & 0.723 & 0.693 \\
&& ViT       & base & ImageNet-1k & 0.657 & 0.646 & 0.657 & 0.657 \\
&& ViT       & base & ImageNet-21k & \textbf{0.928} & \textbf{0.928} & \textbf{0.930} & \textbf{0.928} \\
&& ViT       & base & None         & 0.272 & 0.212 & 0.262 & 0.272 \\
&& ViT       & small & ImageNet-1k & 0.720 & 0.713 & 0.716 & 0.720 \\
&& ViT       & small & None         & 0.188 & 0.090 & 0.154 & 0.188 \\
\midrule
\multirow{9}{*}{A} & \multirow{9}{*}{SUR} 
& ResNet50 & base & ImageNet-1k & \textbf{0.796} & \textbf{0.798} & \textbf{0.804} & \textbf{0.796} \\
&& ResNet50 & base & None         & 0.583 & 0.552 & 0.557 & 0.583 \\
&& ResNet152 & base & ImageNet-1k & 0.742 & 0.751 & 0.774 & 0.742 \\
&& ResNet152 & base & None         & 0.506 & 0.491 & 0.521 & 0.506 \\
&& ViT       & base & ImageNet-1k & 0.492 & 0.472 & 0.478 & 0.492 \\
&& ViT       & base & ImageNet-21k & 0.768 & 0.769 & 0.783 & 0.769 \\
&& ViT       & base & None         & 0.230 & 0.153 & 0.135 & 0.230 \\
&& ViT       & small & ImageNet-1k & 0.554 & 0.528 & 0.522 & 0.554 \\
&& ViT       & small & None         & 0.235 & 0.158 & 0.172 & 0.235 \\
\midrule
\multirow{9}{*}{B} & \multirow{9}{*}{MIX} 
& ResNet50 & base & ImageNet-1k & 0.850 & 0.851 & 0.858 & 0.850 \\
&& ResNet50 & base & None         & 0.772 & 0.764 & 0.795 & 0.772 \\
&& ResNet152 & base & ImageNet-1k & 0.774 & 0.773 & 0.809 & 0.774 \\
&& ResNet152 & base & None         & 0.705 & 0.701 & 0.728 & 0.705 \\
&& ViT       & base & ImageNet-1k & 0.769 & 0.760 & 0.788 & 0.769 \\
&& ViT       & base & ImageNet-21k & \textbf{0.883} & \textbf{0.883} & \textbf{0.890} & \textbf{0.883} \\
&& ViT       & base & None         & 0.505 & 0.472 & 0.504 & 0.505 \\
&& ViT       & small & ImageNet-1k & 0.700 & 0.691 & 0.742 & 0.700 \\
&& ViT       & small & None         & 0.410 & 0.325 & 0.446 & 0.410 \\
\midrule
\multirow{9}{*}{B} & \multirow{9}{*}{SEC} 
& ResNet50 & base & ImageNet-1k & 0.645 & 0.593 & 0.678 & 0.645 \\
&& ResNet50 & base & None         & 0.773 & 0.760 & 0.783 & 0.773 \\
&& ResNet152 & base & ImageNet-1k & 0.648 & 0.615 & 0.712 & 0.648 \\
&& ResNet152 & base & None         & 0.724 & 0.701 & 0.737 & 0.724 \\
&& ViT       & base & ImageNet-1k & 0.739 & 0.733 & 0.745 & 0.739 \\
&& ViT       & base & ImageNet-21k & \textbf{0.957} & \textbf{0.958} & \textbf{0.961} & \textbf{0.958} \\
&& ViT       & base & None         & 0.289 & 0.217 & 0.306 & 0.289 \\
&& ViT       & small & ImageNet-1k & 0.728 & 0.725 & 0.750 & 0.728 \\
&& ViT       & small & None         & 0.449 & 0.365 & 0.535 & 0.449 \\
\midrule
\multirow{9}{*}{B} & \multirow{9}{*}{SUR} 
& ResNet50 & base & ImageNet-1k & 0.864 & 0.862 & 0.870 & 0.864 \\
&& ResNet50 & base & None         & 0.685 & 0.659 & 0.697 & 0.685 \\
&& ResNet152 & base & ImageNet-1k & 0.759 & 0.752 & 0.795 & 0.759 \\
&& ResNet152 & base & None         & 0.660 & 0.614 & 0.591 & 0.660 \\
&& ViT       & base & ImageNet-1k & 0.704 & 0.700 & 0.730 & 0.704 \\
&& ViT       & base & ImageNet-21k & \textbf{0.888} & \textbf{0.891} & \textbf{0.900} & \textbf{0.891} \\
&& ViT       & base & None         & 0.429 & 0.323 & 0.305 & 0.429 \\
&& ViT       & small & ImageNet-1k & 0.706 & 0.703 & 0.721 & 0.706 \\
&& ViT       & small & None         & 0.428 & 0.367 & 0.489 & 0.428 \\
\bottomrule
\end{tabular}
}
\end{table*}

\section{Qualitave Peformance Analyses}\label{sec:feature_visualization}

\subsection{Feature Visualization}

\begin{figure}[h!]
  \centering
  \subfloat[Dataset A. Ex-vivo CCD Camera images.]{%
    \includegraphics[width=1\textwidth]{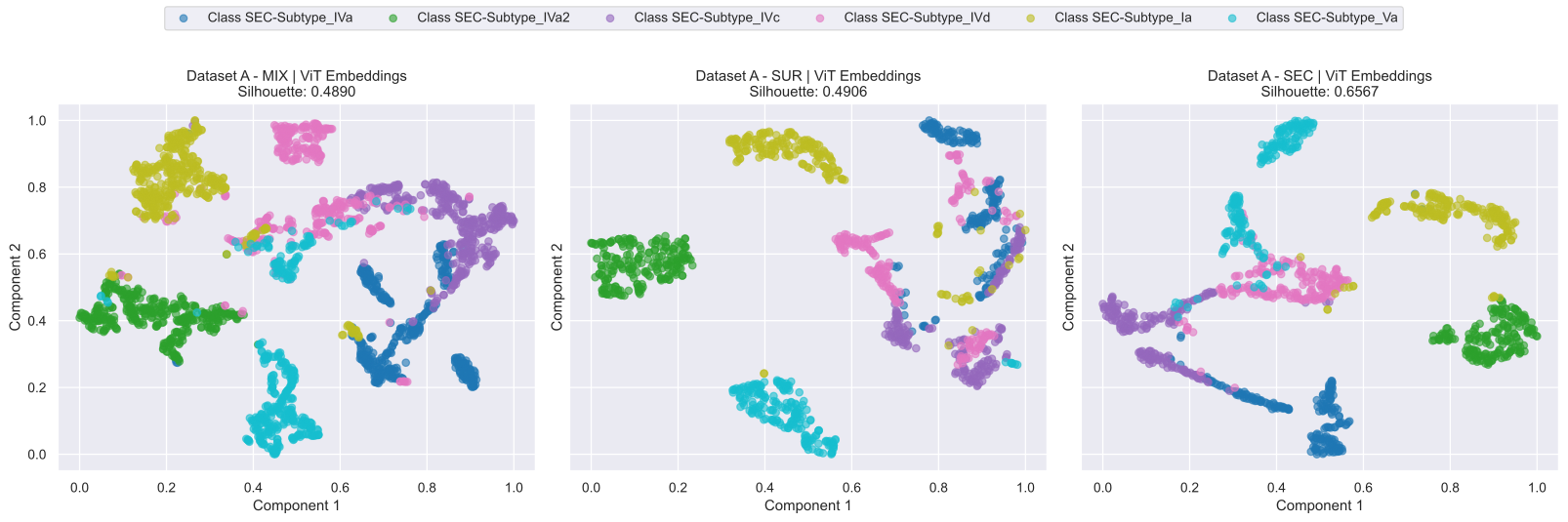}%
    \label{fig:datasetA}
  }
  \vspace{1cm} 
  \subfloat[Dataset B. Ex-vivo endoscopic images.]{%
    \includegraphics[width=1\textwidth]{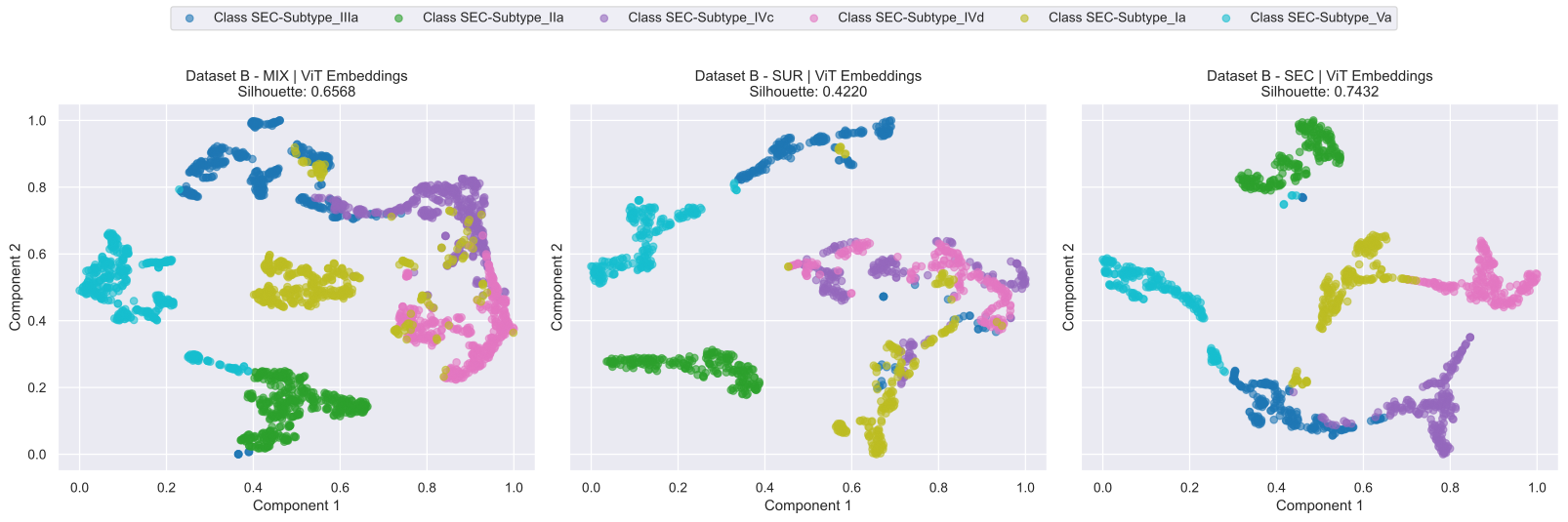}%
    \label{fig:datasetB}
  }
  \caption{2D t-SNE visualization of ViT feature embeddings extracted from the test sets of Dataset A (CCD-Camera images) and Dataset B (Endoscopic images) across their respective subsets. Each point represents a kidney stone sample, with colors indicating different subtypes. The silhouette score is shown in the title of each subset plot, reflecting how well the samples are clustered: higher scores indicate better-defined and more distinct groupings among subtypes.}
  \label{fig:datasets_comparison}
\end{figure}

Understanding the internal representations of the ViT model is crucial for interpreting its decision-making process. Extracting and visualizing the embeddings generated by the model (see Figure~\ref{fig:datasets_comparison}) provides insights into how it differentiates between various classes. This section outlines the procedure for obtaining embeddings from a pre-trained ViT model and visualizing them using t-Distributed Stochastic Neighbor Embedding (t-SNE).

The ViT model processes an input image \( \mathbf{I} \) of dimensions \( H \times W \times C \), where \( H \) and \( W \) denote the height and width, and \( C \) represents the number of color channels (e.g., 3 for RGB images). 

The extraction of embeddings involves the following steps:

\begin{enumerate}
    \item \textbf{Dividing the Image into Patches:} Split the image \( \mathbf{I} \) into a grid of non-overlapping patches, each of size \( P \times P \). This results in \( N = \frac{HW}{P^2} \) patches.

    \item \textbf{Flattening and Projecting Patches:} Flatten each patch into a vector and project it into a \( D \)-dimensional embedding space using a linear transformation. This produces a set of patch embeddings \( \{\mathbf{z}_1, \mathbf{z}_2, \ldots, \mathbf{z}_N\} \), where each \( \mathbf{z}_i \in \mathbb{R}^D \).

    \item \textbf{Adding Positional Information:} Since the Transformer architecture does not inherently capture the spatial arrangement of patches, add positional encodings to each patch embedding to retain information about their original positions within the image.

    \item \textbf{Processing Through Transformer Layers:} Pass the positionally encoded patch embeddings through multiple Transformer layers, which apply self-attention mechanisms to capture relationships between patches. The output after these layers provides the final embeddings for each patch.

    \item \textbf{Extracting the [CLS] Token Embedding:} In many ViT architectures, a special classification token ([CLS]) is added to the sequence of patch embeddings. The output corresponding to this token after the Transformer layers, denoted as \( \mathbf{z}_{\text{CLS}} \in \mathbb{R}^D \), serves as a comprehensive representation of the entire image and can be used for tasks like classification or visualization.

    \item \textbf{Visualizing Feature Embeddings with t-SNE:} To interpret and analyze the high-dimensional [CLS] token embeddings (\( \mathbf{z}_{\text{CLS}} \)), we apply t-SNE, a nonlinear dimensionality reduction technique. By emphasizing the preservation of local structures, t-SNE effectively uncovers clusters and nonlinear patterns within the data, making it an ideal choice for visualizing the complex embeddings produced by the ViT model.

    \item \textbf{Evaluating Cluster Quality with the Silhouette Score:} To quantitatively assess the quality of the clusters formed in the 2D t-SNE visualization, compute the Silhouette Score. This metric measures how similar each data point is to its own cluster compared to other clusters, with values ranging from -1 to 1. A higher Silhouette Score indicates that data points are well-matched to their own cluster and distinctly separated from other clusters, suggesting a meaningful and well-defined clustering structure. This evaluation helps validate the effectiveness of the extracted feature embeddings and the appropriateness of the t-SNE visualization.
\end{enumerate}

\subsection{Comparing ViT and CNN Feature Extraction}\label{sec:vit_vs_cnn}

In this section, a comparative analysis of feature embeddings extracted using ViT and CNN models is presented. The focus is placed on Dataset B, specifically the MIX subset, to evaluate how each model represents kidney stone images. MIX was selected because it simultaneously captures the visual characteristics of both SUR and SEC views of kidney stones. For the CNN-based approach, the ResNet50 architecture is employed, trained under the same experimental conditions used for the ViT models to ensure a fair comparison. This analysis aims to highlight the differences in how convolutional and transformer-based architectures encode visual information.

\begin{figure}[h!]
  \centering
  \includegraphics[width=1\textwidth]{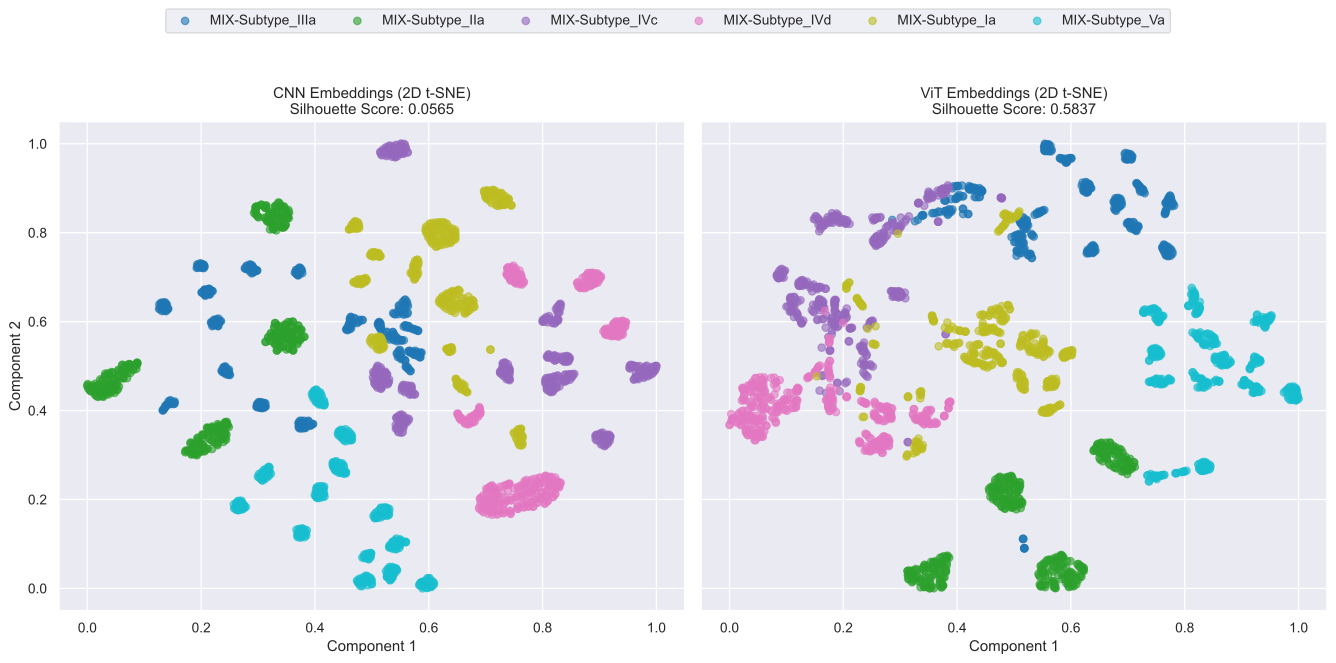}
  \caption{Comparison of 2D visualizations of feature embeddings from Dataset B (MIX subset) testing dataset, extracted using the CNN (left) and ViT (right) models. Each point represents a kidney stone sample, with colors indicating different subtypes. The silhouette score is displayed in the title of each plot; higher values indicate better-defined and more cohesive clustering of the subtypes.
}
  \label{fig:cnn_vs_vit_mix}
\end{figure}

The comparison reveals that the ViT model produces embeddings (see Figure \ref{fig:cnn_vs_vit_mix}) with well-defined clusters and higher silhouette scores, indicating its ability to capture more discriminative features that effectively differentiate between kidney stone subtypes. In contrast, the CNN  model, while still able to form clusters, yields less distinct groupings with lower silhouette scores, suggesting it captures less discriminative features for this particular classification task.

\subsubsection{Intra-Class Distance Analysis}

The average intra-class distance measures how closely the feature representations (embeddings) of items within the same class are grouped together. Smaller values indicate that the embeddings are more tightly clustered, which is beneficial for classification tasks. Across all kidney stone subtypes in Dataset B MIX subset, the ViT consistently shows substantially lower intra-class distances compared to the CNN (see Table \ref{table:class_distance}). This suggests that the ViT model produces more compact and consistent feature representations within each class.

\begin{table}[]
\centering
\caption{Average Intra-Class Distances for CNN and ViT Models on Dataset B (MIX subset). Lower values in bold are better, indicating more compact embeddings.}
\begin{tabular}{lcc}
\hline
\textbf{Class}   & \textbf{CNN Distance} & \textbf{ViT Distance} \\ \hline
MIX-Subtype IIIa & 10.8954               & \textbf{2.9526}       \\
MIX-Subtype IIa  & 6.2529                & \textbf{0.7863}       \\
MIX-Subtype IVc  & 12.4632               & \textbf{4.8948}       \\
MIX-Subtype IVd  & 10.9059               & \textbf{2.7534}       \\
MIX-Subtype Ia   & 12.3673               & \textbf{5.9586}       \\
MIX-Subtype Va   & 9.7938                & \textbf{3.1757}       \\ \hline
\end{tabular}
\label{table:class_distance}
\end{table}

This indicates that the ViT model produces more compact and coherent feature representations for each class. Specifically, for MIX-Subtype IIa, the ViT model reduces the average intra-class distance from 6.2529 (CNN) to 0.7863, highlighting a significant improvement in embedding compactness. These findings suggest that the ViT effectively captures global contextual information through self-attention mechanisms, leads to more discriminative and structured feature spaces.

\begin{figure}[h!]
  \centering
  \includegraphics[width=1\textwidth]{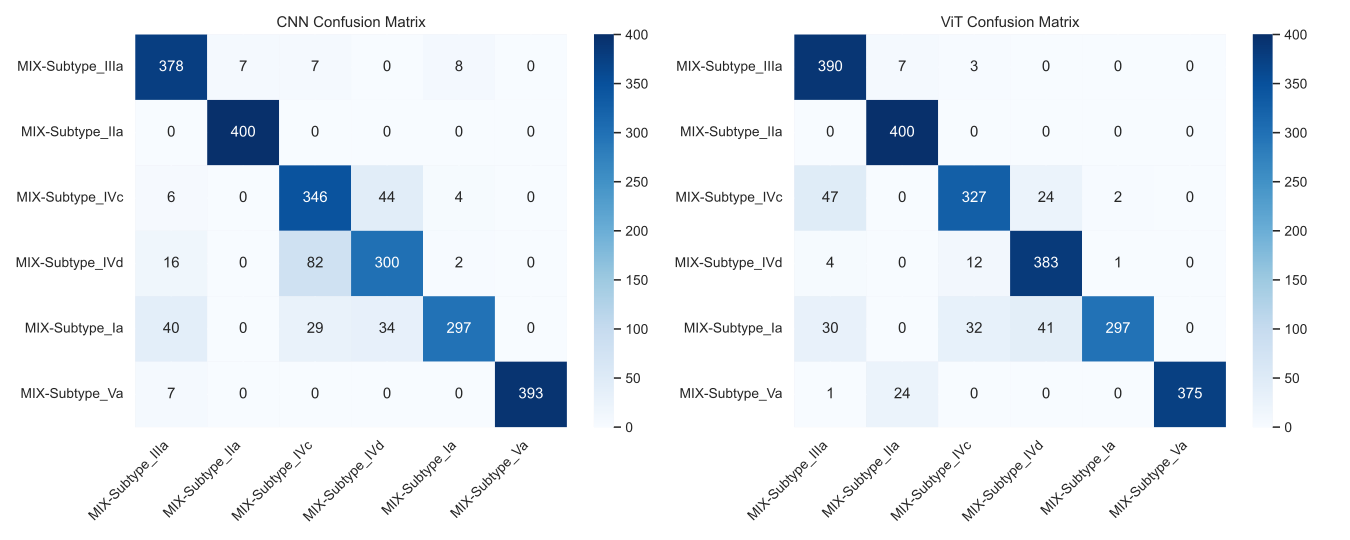}
    \caption{Comparison of confusion matrices for the CNN (left) and ViT (right) models on the Dataset B (MIX subset). The ViT model shows improved classification performance with higher correct predictions and fewer misclassifications, particularly for challenging subtypes such as MIX-Subtype IVd.}
  \label{fig:confusion_matrix}
\end{figure}

\subsubsection{Per-Class Performance Analysis}

A comparison of the confusion matrices (see Figure \ref{fig:confusion_matrix}) for the CNN (left) and ViT (right) models reveals notable differences in their classification performance. For example, in the MIX-Subtype IVd category, the CNN correctly classified only 300 samples with 100 misclassifications, whereas the ViT achieved 383 correct predictions with merely 17 misclassifications. Additionally, while both models performed similarly for MIX-Subtype IIa and MIX-Subtype Ia, the ViT model also improved the correct classification of MIX-Subtype IIIa (390 versus 378 for CNN). Although CNN shows a slight advantage in MIX-Subtype IVc and MIX-Subtype Va, the overall pattern suggests that the ViT model yields more robust and discriminative feature representations. This is particularly evident in its ability to better differentiate challenging subtypes such as MIX-Subtype IVd, resulting in fewer critical misclassifications and enhanced overall clustering of the kidney stone subtypes.

\begin{table}[]
\centering
\caption{Per-Class Performance Metrics Comparison between CNN and ViT Models on Dataset B (MIX subset). Best results are highlighted in bold.}
\begin{tabular}{lcccccccc}
\hline
\multirow{2}{*}{Class} & \multicolumn{2}{c}{Accuracy} & \multicolumn{2}{c}{Precision} & \multicolumn{2}{c}{Recall} & \multicolumn{2}{c}{F1 Score} \\ \cline{2-9} 
                 & CNN            & ViT            & CNN            & ViT            & CNN            & ViT            & CNN            & ViT            \\ \hline
MIX-Subtype IIIa & 0.962          & 0.962          & \textbf{0.846} & 0.826          & 0.945          & \textbf{0.975} & 0.893          & \textbf{0.894} \\
MIX-Subtype IIa  & \textbf{0.997} & 0.987          & \textbf{0.983} & 0.928          & 1.000          & 1.000          & \textbf{0.991} & 0.963          \\
MIX-Subtype IVc  & 0.928          & \textbf{0.950} & 0.746          & \textbf{0.874} & \textbf{0.865} & 0.818          & 0.801          & \textbf{0.845} \\
MIX-Subtype IVd  & 0.926          & \textbf{0.966} & 0.794          & \textbf{0.855} & 0.750          & \textbf{0.958} & 0.771          & \textbf{0.903} \\
MIX-Subtype Ia   & 0.951          & \textbf{0.956} & 0.955          & \textbf{0.990} & 0.743          & 0.743          & 0.835          & \textbf{0.849} \\
MIX-Subtype Va   & \textbf{0.997} & 0.990          & 1.000          & 1.000          & \textbf{0.983} & 0.938          & \textbf{0.991} & 0.968          \\ \hline
\end{tabular}%
\label{table:perclass}
\end{table}

Complementing the confusion--matrix insights, the per--class statistics in Table \ref{table:perclass} confirm that the ViT consistently translates its cleaner decision boundaries into stronger quantitative performance. The transformer either matches or exceeds the CNN’s accuracy in every subtype, posting clear gains in the more challenging MIX-Subtype IVc (+2.2\%) and IVd (+4.0\%). These margins widen when \emph{recall} and \emph{F1} are considered: ViT boosts recall by more than 20\% in IVd and secures the top F1 in four of the six categories, underscoring its ability to retrieve a larger proportion of true positives while keeping errors low. 

Although the CNN model exhibits higher precision in certain classes such as MIX-Subtype IIIa and MIX-Subtype IIa, suggesting lower false‐positive rates, its recall is notably lower in more complex subtypes. This contrast highlights the strength of the ViT in capturing global contextual information through self‐attention mechanisms, enabling superior performance in nuanced classification tasks and yielding more balanced, subtype‐specific metrics.

Taken together, these findings indicate that the ViT architecture delivers the most reliable and comprehensive performance across Dataset B MIX subset, making it the preferred choice for robust kidney‐stone subtype classification.

\subsection{Qualitative Analysis of CNN and ViT}

\begin{figure}[h!]
  \centering
  \includegraphics[width=1\textwidth]{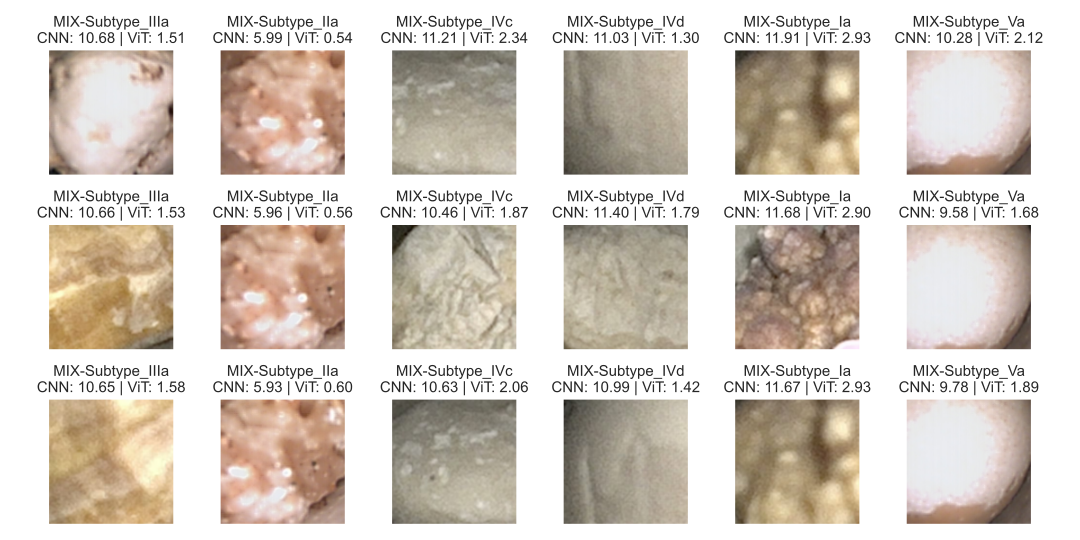}
\caption{Qualitative comparison of test samples from Dataset B (MIX subset) that are poorly modeled by the CNN but well represented by the ViT model. Each column corresponds to a specific class, and each row displays samples with the highest difference in distance to the class center between CNN and ViT embeddings. The reported CNN and ViT distances indicate the Euclidean distance to the corresponding class center; lower values represent more compact and consistent embeddings. }
  \label{fig:qualitative}
\end{figure}

To further investigate the representational capabilities of the ViT and CNN models, we perform a qualitative analysis of selected test samples from Dataset B (MIX subset). Specifically, we focus on samples that are poorly modeled by the CNN but well captured by the ViT model. For each class, we identify samples where the CNN embedding is far from its corresponding class center, while the ViT embedding is close. The Euclidean distance between a sample's embedding and the class centroid is used to quantify this proximity. 

Samples with the largest values are selected, as they represent cases where the CNN embedding is significantly less compact compared to the ViT embedding. Figure \ref{fig:qualitative}  displays the selected samples, arranged such that each column corresponds to a kidney stone subtype and each row presents examples with the largest discriminative gap. Each image is annotated with its CNN and ViT embedding distances.

This qualitative comparison illustrates that ViT produces more tightly clustered and representative embeddings across multiple classes, particularly in complex and visually ambiguous subtypes. These observations are consistent with the quantitative metrics and further support the superiority of ViT for robust feature extraction in the context of kidney stone classification.

\section{Conclusions}\label{sec:conclusion}

This study presents a ViT-based approach for kidney stone classification, demonstrating superior performance compared to conventional CNN methods in terms of classification accuracy and feature representation. The proposed ViT architecture yields more compact and discriminative embeddings, as evidenced by lower intra-class distances and clearer clustering of kidney stone subtypes. These advantages are attributed to the self-attention mechanism, which effectively captures global contextual information.

To further assess robustness, future work will incorporate synthetic image corruptions during testing to evaluate the stability of the learned embeddings under perturbations. This analysis aims to provide deeper insights into the resilience of the approach and its potential for reliable deployment in real-world clinical environments. Overall, the findings underscore the effectiveness of ViT-based architectures as promising and robust tools for automated kidney stone classification.

\section*{Acknowledgments}
The authors wish to acknowledge the Mexican Secretaría de Ciencia, Humanidades, Tecnología e Innovación (Secihti) for their support in terms of postgraduate scholarships in this project, and the Data Science Hub at Tecnologico de Monterrey for their support on this project.
This work has been supported by Azure Sponsorship credits granted by Microsoft's AI for Good Research Lab through the AI for Health program.
The project was also supported by the French-Mexican ANUIES CONAHCYT Ecos Nord grant 322537.
We also gratefully acknowledge the support from the Google Explore Computer Science Research (CSR) Program for partially funding this project through the LATAM Undergraduate Research Program.

{\small
\bibliographystyle{splncs04}
\bibliography{biblio}
}

\end{document}